\documentclass[sigconf,review=false]{acmart}

\usepackage{siunitx}
\usepackage{pifont}% http://ctan.org/pkg/pifont
\usepackage{layouts}
\usepackage{bm}

\usepackage{booktabs}
\usepackage{multirow}
\usepackage{microtype}
\usepackage{enumitem}

\renewcommand{\bar}{\overline}

\newcommand{\cmark}{\ding{51}}%
\newcommand{\xmark}{\ding{55}}%

\usepackage{threeparttable}

\usepackage{stmaryrd}

\usepackage{xspace}

\usepackage{algorithm}[H] % Use algorithm but do not float
\usepackage[noend]{algpseudocode}% http://ctan.org/pkg/algorithmicx
\newbox\statebox
\newcommand{\myState}[1]{%
    \setbox\statebox=\vbox{#1}%
    \edef\thealgruleheight{\dimexpr \the\ht\statebox+1pt\relax}%
    \edef\thealgruledepth{\dimexpr \the\dp\statebox+1pt\relax}%
    \ifdim\thealgruleheight<.75\baselineskip
        \def\thealgruleheight{\dimexpr .75\baselineskip+1pt\relax}%
    \fi
    \ifdim\thealgruledepth<.25\baselineskip
        \def\thealgruledepth{\dimexpr .25\baselineskip+1pt\relax}%
    \fi
    \State #1%
    \def\thealgruleheight{\dimexpr .75\baselineskip+1pt\relax}%
    \def\thealgruledepth{\dimexpr .25\baselineskip+1pt\relax}%
}

\AtBeginDocument{%
  \providecommand\BibTeX{{%
    \normalfont B\kern-0.5em{\scshape i\kern-0.25em b}\kern-0.8em\TeX}}}

\setcopyright{acmcopyright}
\copyrightyear{2023}
\acmYear{2023}
\setcopyright{rightsretained}
\acmConference[GECCO '23]{Genetic and Evolutionary Computation Conference}{July 15--19, 2023}{Lisbon, Portugal}
\acmBooktitle{Genetic and Evolutionary Computation Conference (GECCO '23), July 15--19, 2023, Lisbon, Portugal}
\acmDOI{10.1145/3583131.3590498} \acmISBN{979-8-4007-0119-1/23/07}

\newcommand{\cD}{\mathcal{D}}

\newcommand{\cN}{\mathcal{N}}

\newcommand{\cP}{\mathcal{P}}

\renewcommand{\vec}[1]{{\bm{{#1}}}} % vector
\newcommand{\mat}[1]{{\bm{{#1}}}} % matrix

\newcommand{\iI}{\mat I}

\newcommand{\params}{\vec{\theta}}

\newcommand{\fitness}{f}

\newcommand{\expect}[2]{ \mbox{E}_{ {#1} } \left[ {#2} \right]}
\newcommand{\var}[2]{ \mbox{Var}_{ {#1} } \left[ {#2} \right]}
\newcommand{\proba}[1]{\mbox{P} \left( {#1} \right)}

\newcommand{\cell}{\mathcal{C}}

\newcommand{\name}{\textsc{ARIA}\xspace}

\newcommand{\pga}{\textsc{PGA}\xspace}

\newcommand{\meinit}{\textsc{ME init}\xspace}

\newcommand{\esinit}{\textsc{ES init}\xspace}
\newcommand{\pgainit}{\textsc{PGA init}\xspace}

\newcommand{\nameME}{\textsc{\name - ME init}\xspace}
\newcommand{\nameES}{\textsc{\name - ES init}\xspace}
\newcommand{\namePGA}{\textsc{\name - PGA init}\xspace}

\newcommand{\linearrim}{\textsc{Linear RIM}\xspace}

\newcommand{\mapelites}{\textsc{MAP-Elites}\xspace}

\newcommand{\pgame}{PGA-ME\xspace}

\newcommand{\centroid}{\vec c}
\newcommand{\bd}{\vec d}

\newcommand{\normeucl}[1]{\left\lVert#1\right\rVert_2}

\newcommand{\distDesc}{\cD_{d}}
\newcommand{\distFit}{\cD_{f}}

\newcommand{\distDescParam}{\distDesc^{(\params)}}
\newcommand{\distFitParam}{\distFit^{(\params)}}

\newcommand{\distDescParamI}{\distDesc^{(\params_i)}}
\newcommand{\distFitParamI}{\distFit^{(\params_i)}}

\newcommand{\obj}{o}
\newcommand{\noisyobj}{\widetilde{\obj}}

\newcommand{\meanopt}{\vec \phi}
\newcommand{\std}{\sigma}
\newcommand{\utility}{u}

\newcommand{\noise}{\vec \epsilon}

  \newcommand{\meanoptInit}{\meanopt_{\mbox{init}}}
  \newcommand{\numsamples}{N_{\mbox{s}}}
  \newcommand{\numreeval}{N_r}
  \newcommand{\numGradientAscent}{N_{\mbox{grad}}}
  \newcommand{\gradientEstimate}{\mbox{gradient\_estimate}}

\newcommand{\numReplications}{M}
\newcommand{\numCells}{ N_{\cell} }

\newcommand{\iverson}[1]{ \varphi \left( {#1} \right)  }

\newcommand{\reprodFunc}{{\textsc{Reproducibility Improvement}}\xspace}

\newcommand{\explored}{\cP_{\mbox{expl}}}
\newcommand{\togo}{\cP_{\mbox{targ}}}

\newcommand{\celltogo}{\cell_{\mbox{targ}}}
\newcommand{\meanbd}{\bar{\bd}}

\newcommand{\brax}{\textsc{Brax}\xspace}
\newcommand{\qdax}{\textsc{QDax}\xspace}

\newcommand{\nvs}{NDV\xspace}
\newcommand{\ef}{EF\xspace}

\newcommand{\momer}{\textsc{MOME-R}\xspace}
\newcommand{\mome}{\textsc{MOME}\xspace}
\newcommand{\mesa}{\textsc{ME-Sa}\xspace}
\newcommand{\mesar}{\textsc{ME-Sa-R}\xspace}

\newcommand{\nameMEshort}{\name-ME\xspace}

\newcommand{\cellstart}{\cell_{\mbox{s}}}

\newcommand{\meanopttogo}{\meanopt_{\mbox{targ}}}

\newtheorem{theorem}{Theorem}

\newtheorem{lemma}{Lemma}

\makeatletter
\newcommand\blfootnote[1]{%
  \begingroup
  \renewcommand{\@makefntext}[1]{\noindent\makebox[1.8em][r]#1}
  \renewcommand\thefootnote{}\footnote{#1}%
  \addtocounter{footnote}{-1}%
  \endgroup
}
\makeatother

\begin{document}

% \title{Optimizing for Behavioral Reproducibility in Massively Parallelizable Quality-Diversity Tasks}
% \title{Optimizing Explicitly for Behavioral Reproducibility\\in Uncertain Quality-Diversity Tasks}
\title{Don't Bet on Luck Alone: Enhancing Behavioral Reproducibility of Quality-Diversity Solutions in Uncertain Domains}
% With Evolution Strategies?
% High-dimensional controllers?

\author{Luca Grillotti}
\authornote{Both authors contributed equally to this research.}
\orcid{0000-0003-4539-8211}
\affiliation{%
% \department{Adaptive and Intelligent Robotics Lab}
\institution{Imperial College London}
\city{London}
\country{United Kingdom}
}
\email{luca.grillotti16@imperial.ac.uk}

\author{Manon Flageat}
\authornotemark[1]
\orcid{0000-0002-4601-2176}
\affiliation{%
% \department{Adaptive and Intelligent Robotics Lab}
\institution{Imperial College London}
\city{London}
\country{United Kingdom}
}
\email{manon.flageat18@imperial.ac.uk}

\author{Bryan Lim}
\orcid{0000-0002-2324-1400}
\affiliation{%
% \department{Adaptive and Intelligent Robotics Lab}
\institution{Imperial College London}
\city{London}
\country{United Kingdom}
}
\email{bryan.lim16@imperial.ac.uk}

\author{Antoine Cully}
\orcid{0000-0002-3190-7073}
\affiliation{%
% \department{Adaptive and Intelligent Robotics Lab}
\institution{Imperial College London}
\city{London}
\country{United Kingdom}
}
\email{a.cully@imperial.ac.uk}

\renewcommand{\shortauthors}{Grillotti et al.}


\begin{abstract}
Quality-Diversity (QD) algorithms are designed to generate collections of high-performing solutions while maximizing their diversity in a given descriptor space.
However, in the presence of unpredictable noise, the fitness and descriptor of the same solution can differ significantly from one evaluation to another, leading to uncertainty in the estimation of such values.
Given the elitist nature of QD algorithms, they commonly end up with many degenerate solutions in such noisy settings.
In this work, we introduce Archive Reproducibility Improvement Algorithm (ARIA); a plug-and-play approach that improves the reproducibility of the solutions present in an archive.
We propose it as a separate optimization module, relying on natural evolution strategies, that can be executed on top of any QD algorithm. 
Our module mutates solutions to (1) optimize their probability of belonging to their niche, and (2) maximize their fitness.
The performance of our method is evaluated on various tasks, including a classical optimization problem and two high-dimensional control tasks in simulated robotic environments.
We show that our algorithm enhances the quality and descriptor space coverage of any given archive by at least 50\%.
\end{abstract}

\begin{CCSXML}
<ccs2012>
<concept>
<concept_id>10010147.10010178.10010213.10010204.10011814</concept_id>
<concept_desc>Computing methodologies~Evolutionary robotics</concept_desc>
<concept_significance>500</concept_significance>
</concept>
</ccs2012>
\end{CCSXML}

\ccsdesc[500]{Computing methodologies~Evolutionary robotics}

\keywords{Quality-Diversity, Uncertain domains, Neuroevolution}

\begin{teaserfigure}
\centering
  \vspace{-15pt}\includegraphics[width=0.90\textwidth]{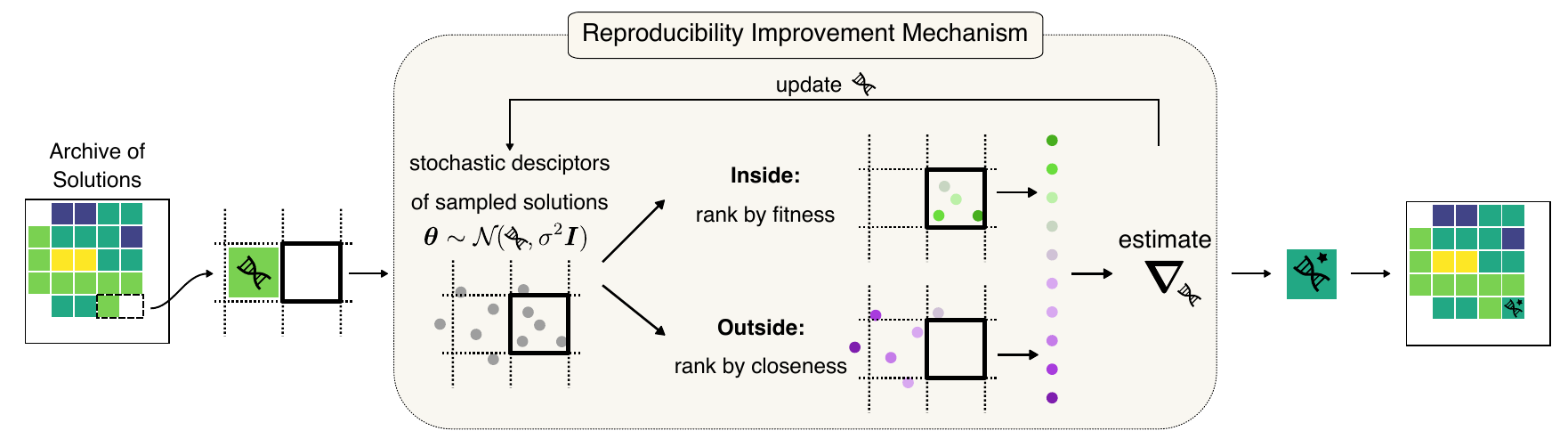}
  \vspace{-10pt}\caption{One iteration of the \textit{Archive Reproducibility Improvement Algorithm} (\name{}) during the "completion phase".
  Two adjacent cells are selected such that one, the target cell, is empty while the other one is already populated by a solution.
  Then, the \textit{Reproducibility Improvement Mechanism} uses this solution as a starting point for finding a high-performing solution in the target cell.
  After several steps of Reproducibility Improvement, we add the resulting optimized solution to the target cell.
  }
  \label{fig:teaser:aria}
\end{teaserfigure}

\maketitle

\vspace{-5pt} \section{Introduction}

\begin{figure*}[t]
    \centering
    \includegraphics[width=0.70\textwidth]{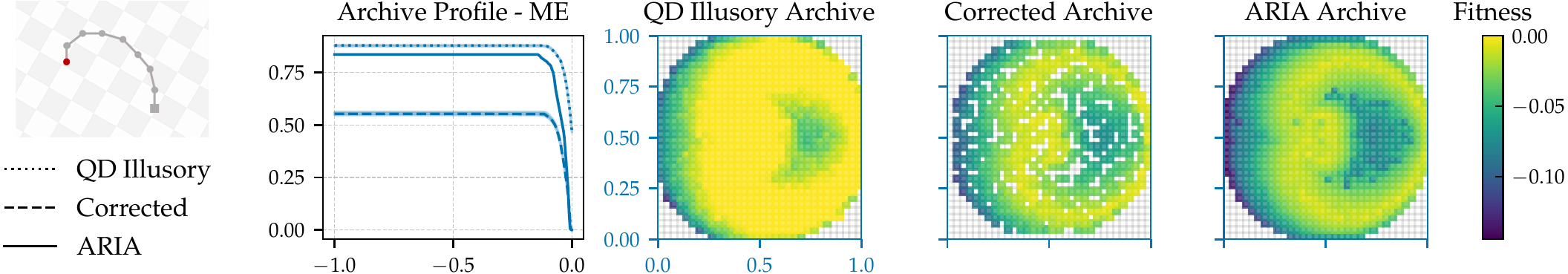}
    \vspace{-10pt}\caption{Archive Profile, Illusory archive, and the corresponding corrected archive returned by \nameME on the Arm task. The ARIA archive shows the result obtained after optimization from \name.}
    \label{fig:aria-improvement-over-lucky}
\end{figure*}

Quality-Diversity (QD) Algorithms~\cite{Pugh2015, Pugh2016, Cully2018} are evolutionary algorithms designed to produce collections, or \textit{archives}, of diverse and high-performing solutions. 
These archives can then be used for various applications, including damage recovery in robotics~\cite{Cully2014, allard2022hierarchicalHTE, kaushik2020adaptivepriorAPROL,chatzilygeroudis2018reset}, video-game design~\cite{fontaine2020illuminating, earle2022illuminating}, and configuring urban layouts~\cite{galanos2021archurbandesign}.

QD approaches rely on an evaluation process which is used to estimate the performance of a given solution; for each solution an evaluation also results in a descriptor which characterizes the novelty of that solution compared to the others.
Most QD algorithms are designed under the assumption that this evaluation process is deterministic.
Deterministic here means that the evaluation of a solution always returns the same result (fitness and descriptor).
However, realistically, there are many contexts and applications in which uncertainty is present in the evaluation process.
For example, in robotics, the measurements from a sensor may be noisy and applying a command to an actuator may not lead to the exact desired displacement; in those cases, sending a command to a robot may not always return the same results.
QD setups where the domain presents uncertainty belong to the Uncertain QD framework~\cite{flageat2023uncertain}.
If QD algorithms are not designed to take this uncertainty into account, they may produce "illusory" archives, with many solutions whose performance and descriptors are wrongly estimated and may exhibit a high variance~\cite{flageat2023uncertain}.
In such cases, we say that the solutions have low \textit{reproducibility}.
To have highly reproducible solutions, they need to have a low variance and good descriptor and performance estimations.
%If we want to be able to predict which descriptor a solution will have, we expect that solution to be highly reproducible.

In this work, we introduce the \textit{Archive Reproducibility Improvement Algorithm} (\name) a method that lowers down the solution's variance, i.e. increases their reproducibility.
It can be used as a plug-and-play tool, applicable to any set of solutions, including solutions returned by QD and Evolution Strategy algorithms.
It relies on a Reproducibility Improvement Mechanism (Fig.~\ref{fig:teaser:aria}) which uses natural evolution strategies to decrease the descriptor variance of a solution and also maximize its fitness.
We leverage this mechanism to not only increase the reproducibility of solutions present in an archive, but also to discover new solutions, hence increasing the diversity of a given archive (Fig.~\ref{fig:aria-improvement-over-lucky}).
We compare \name with different reproducibility-aware baselines on three classical Uncertain QD tasks, and show that \name finds the best archives, both in terms of performance and reproducibility.
More specifically, compared to standard baselines, the best variants of \name achieve a 30\% higher QD-Score, and improve the reproducibility by at least 20\%.

\section{Background}

\subsection{Evolution Strategies}

\label{sec:nes}

In this work, we intend to use ES optimization algorithms that are applicable to possibly high-dimensional problems, including deep neuroevolution tasks.
\citet{ESScalableAlternative} proposes a method that takes inspiration from Natural Evolution Strategies \cite{NESWierstra}: instead of optimizing directly for an objective function $\noisyobj(\cdot)$, they maximize the expectation of $\noisyobj$ under a Gaussian distribution: $\expect{\params\sim\cN(\meanopt,\std^2 \iI)}{\noisyobj(\params)}$
where the mean $\meanopt$ is the variable to optimize, and the standard deviation $\std$ is kept constant.
%
% The gradient of such expectation can be estimated as in REINFORCE~\cite{williams1992simple}:
% %
% \begin{align}
%     \nabla_\meanopt \expect{\params\sim\cN(\meanopt,\std^2 \iI)}{\noisyobj(\params)} 
%     % &= \expect{ \params\sim\cN(\meanopt,\std^2 \iI) } { \noisyobj(\params) \nabla_{\meanopt}\cN(\meanopt,\std^2 \iI) } \\
%     % &= \frac{1}{\sigma} \expect{ \noise \sim\cN(\vec 0, \iI) }{ \noisyobj(\meanopt + \sigma \noise)\noise } \\
%     &\approx \frac{1}{\sigma} \sum_{s=1}^{\numsamples} \noisyobj(\meanopt + \sigma \noise_s) \noise_s \quad \noise_s\sim \cN(\vec 0, \iI)
% \end{align}
%
It is common in ES to use rank-based utility values $(\utility_s)$ instead of the objective values~\cite{NESWierstra, hansen2016cmaestutorial}.
In that case, the gradient used to update the parameters can be expressed: $\frac{1}{\sigma} \sum_{s=1}^{\numsamples} \utility_s \noise_s$, where $\noise_s$ is sampled from a  normal distribution.
%
% Then, the gradient can be estimated as follow (with abuse of notation):
% \begin{align}
%     \nabla_\meanopt \expect{\params\sim\cN(\meanopt,\std^2 \iI)}{\noisyobj(\params)} &\approx \frac{1}{\sigma} \sum_{s=1}^{\numsamples} \utility_s \noise_s \qquad \noise_s\sim \cN(\vec 0, \iI)
% \end{align}
%
% Finally, all the samples can be mirrored to lower down the variance of this gradient estimator \cite{??}.

\subsection{Quality-Diversity Algorithms}

Quality-Diversity (QD) algorithms \cite{Pugh2015, Cully2018} aim at finding a collection, also called "archive", of diverse and high-performing solutions to an optimization problem.
Unlike common evolutionary algorithms, QD algorithms consider that solutions are not only assigned a fitness, but also a descriptor used to measure their novelty.
Common QD algorithms such as \mapelites~\cite{Mouret2015, Cully2014} discretize this descriptor space into a grid. 
For each cell of this grid, those algorithms intend to find one solution whose descriptor is in the cell, and whose fitness maximizes the performance in that cell.
At each iteration, they operate as follow: solutions are selected from the archive and modified via random variations (e.g. mutations, cross-overs...); then their fitness and descriptors are evaluated, and they are added back to the archive. 
More precisely, to add a solution to the archive, we first check which cell its descriptor belongs to.
If this cell is empty, then the solution is added to the cell; otherwise, if the cell is already occupied by another solution, only the best-performing solution is kept.
% Each iteration of those algorithms follow the same steps: (1) selection from the archive, (2) applying random variations (e.g. mutations, cross-overs...), (3) evaluating the fitness and descriptors, and (4) attempt to add the solutions back to the archive. 
%
With those mechanisms, \mapelites and its variants progressively increase the number of cells filled, while also improving the performance of the solutions stored in the archive.

\subsection{Uncertain QD Setting with Descriptor Reproducibility Maximization}

\label{sec:UQD}

There are environments for which successive evaluations of the same solution may lead to different results.
For example in robotics, this is often due to different initial conditions, noise from the sensors, or looseness in the joints.
In the Uncertain QD Setting, formalized by~\citet{flageat2023uncertain}, successive evaluations of the same solution $\params$ may lead to different fitnesses and descriptors.
In such setting, the fitness $\fitness$ and descriptor $\bd$ resulting from a solution $\params$ can be represented as two distributions $\distFitParam$ and $\distDescParam$.
Considering a descriptor space partitioned into cells $\left( \cell_i \right)$, the problem can be expressed as: for each cell $\cell_i$, we intend to find a solution $\params_i$ maximizing the \textit{expected fitness} $\expect{\fitness_i\sim\distDescParamI}{\fitness_i}$ while the expected descriptor $\expect{\bd_i\sim\distDescParamI}{\bd_i}$ belongs to the the cell $\cell_i$.
Additionally, we aim at minimizing the \textit{variance} of the descriptor components over each descriptor dimension~$j$.
In other words, we consider the following multi-objective constrained optimization problems:
\begin{align}
\label{problem:eq:spread}
\begin{split}
    \text{For each cell $\cell_i$, } & \text{find a solution $\params_i$ that:}\\
    &\mbox{maximize}\: \expect{\fitness_i \sim \distFitParamI}{ \fitness_i } \\
    &\mbox{minimize}\: \var{ \bd_i \sim \distDescParamI }{ \bd_i^{(j)} } \; \forall j \\
    &\mbox{subject to}\:  \expect{\bd_i \sim \distDescParamI }{\bd_i}\in\cell_i
\end{split}
\tag{P1}
\end{align}

\begin{algorithm}[t]
  \small
  \caption{\textsc{\name{}}}
  \label{algo:qdr-main}
  \newcommand\algorithmicitemindent{\hspace*{\algorithmicindent}\hspace*{\algorithmicindent}}

\label{algo:qdr}
  
  \begin{algorithmic}[1]
  \footnotesize
  \State{\textbf{Input:} set of genotypes given by an ES or QD algorithm $\Phi$; number gradient steps $\numGradientAscent$; Number samples per gradient step $\numsamples$; standard deviation samples $\sigma$}

    \State{$\explored \gets \emptyset$, $\togo \gets$ all cells partitioning the descriptor space}

    % \State{$\Phi \gets$ genotypes returned by the Initialization Phase } \label{algo:qdr:line:init}
    
\item[] 

\Comment{\textit{Reproducibility Improvement Phase}}

    \For{ genotype $\meanopt \in \Phi $} \label{algo:qdr:line:reprod:beg}
        \State{$(\bd_s)_{s=1}^{\numsamples} \gets$ evaluate genotype $\numsamples$ times and collect descriptors}

        \State{$\cell_j \gets$ find cell containing mean descriptor $\meanbd$}
    
        \State{$\meanopt_j \gets$ \reprodFunc($\meanoptInit=\meanopt$, $\celltogo=\cell_j$, $\numGradientAscent$, $\numsamples$, $\sigma$)
        } 
        \State{Add $\cell_j$ to $\explored$, and remove it from $\togo$} \label{algo:qdr:line:reprod:end}
    \EndFor 

\item[] 

\Comment{\textit{Archive Completion Phase}}

    \While{$\togo$ is not empty} \label{algo:qdr:line:completion:beg}
        \State{Select 2 adjacent cells such that $\cellstart \in \explored$ and  $\celltogo\in\togo$}

        \State{$\meanopt_{\mbox{s}} \gets$ genotype associated to $\cellstart$}

        \State{$\meanopttogo \gets$ \reprodFunc($\meanoptInit=\meanopt_{\mbox{s}}$, $\celltogo$, $\numGradientAscent$, $\numsamples$, $\sigma$} )

        \State{Add $\meanopttogo$ to the target cell $\celltogo$}

        \State{Add $\celltogo$ to $\explored$, and remove it from $\togo$}

    \EndWhile

\item[]

    \State{\Return all genotypes from all cells}

  \end{algorithmic}
\end{algorithm}

\vspace{-10pt}\subsection{Related Work}

\paragraph{QD in Uncertain Domains}

The issue of QD applied to uncertain domains has incrementally emerged in recent literature following advances in QD toward neuroevolution and complex robotics applications \cite{flageat2022benchmarking}.
When faced with uncertainty, QD algorithms struggle to estimate the  expected quality and novelty of solutions.
% As a result, they do not distinguish good solutions from solutions that have been lucky and evaluated as more performing than they truly are. 
% Due to their elitism, QD algorithms even tend to maintain such lucky solutions, leading to a loss in the quality and diversity of the archive as a whole.
% This problem has recently been formalized under the Uncertain Quality-Diversity setting \cite{}. 
Multiple variants of MAP-Elites have been proposed to address this limitation.
Such variants either rely on sampling solutions multiple times - dynamically \cite{justesen2019adaptivesampling, flageat2023uncertain} or not \cite{Cully2018a} -, or rely on more complex mechanisms based on previous individuals \cite{flageat2020deepgrid, flageat2023uncertain}.
Additionally, gradient-augmented QD approaches such as PGA-MAP-Elites \cite{nilsson2021pgamapelites} and MAP-Elites-ES \cite{colas2020mapeliteses} have also proven promising in tackling the issue of uncertainty. Their informed gradient-based mechanism allows them to produce more reproducible solutions than vanilla QD approaches \cite{flageat2022empiricalPGA}.
However, all these previous works rely on modifying the main MAP-Elites algorithm to directly produce better estimates of performance and encourage reproducible solutions. 
In comparison, our approach ARIA proposes a plug-and-play mechanism that improves the result of any given QD algorithm, even those which do not take into account the uncertainty of the task.  

\paragraph{Mixing QD with ES}

Multiple previous works proposed to integrate ES within MAP-Elites. 
Notably, the CMA-ME algorithm \cite{fontaine2020covariance} proposed to use CMA-ES to generate offspring for MAP-Elites. Similarly to ARIA, CMA-ME optimizes for objectives distinct from the raw task fitness, such as the archive improvement score.
Closer to our work, MAP-Elites-ES \cite{colas2020mapeliteses} uses the Natural ES~\cite{NESWierstra} from~\citet{ESScalableAlternative} to produce offspring within MAP-Elites, optimizing either for the task fitness or the solution novelty. 
These two approaches improve upon MAP-Elites by integrating an ES mechanism within the mutation procedure of QD, while ARIA proposes to apply the ES mechanism as a second step, on top of the chosen QD algorithm. 
Also, the objective function used by \name is different from those employed by CMA-ME and MAP-Elites-ES.
Indeed, the objective function of \name is designed to primarily optimize the reproducibility of solutions.
Furthermore, CMA-ME and MAP-Elites-ES could both be used as initialization algorithms for ARIA.

\paragraph{Previous work in targeted improvement for QD}

Distinct branches of work developed mechanisms to select efficient optimization starting-points in a QD archive.
MAP-Elites-ES \cite{colas2020mapeliteses} relies on a biased parent-selection mechanism; this mechanism selects parents that tend to produce offspring solutions with a higher fitness or novelty. 
Similarly, Go-Explore \cite{ecoffet2021firstreturnthenexplore} solves hard-exploration tasks by selecting relevant cells to explore from.
During ARIA's final phase, its selection mechanism iteratively selects a solution at the border of the discovered descriptor space, and optimizes it to find a new solution in an empty adjacent cell, as illustrated on Figure~\ref{fig:teaser:aria}.
% ARIA also uses a selection mechanism to choose the initial solution for both the ES Reproducibility Improvement and the ES Archive Completion. However, ARIA's selection mechanism does not rely on any performance metric but only on distances in the descriptor space.

% \subsection{Quality-Diversity in Uncertain Domains}

% \subsection{Evolution Strategies and Gradient-based Methods for Improving MAP-Elites}

% \begin{itemize}
%     \item PGA-ME
%     \item ME-ES
%     \item CMA-ME and variants
%     \item Approximating Gradients for Differentiable Quality Diversity in
% Reinforcement Learning
% \item Diversity policy gradient for sample efficient quality-diversity optimization
% \item Scaling Covariance Matrix Adaptation MAP-Annealing to High-Dimensional Controllers
% \item Go-Explore.
% \end{itemize}

\section{Methods}

In this section, we first describe an alternative problem formulation to the Uncertain QD problem~\ref{problem:eq:spread}.
Then, we describe the Archive Reproducibility Improvement Algorithm (\name), a plug-and-play approach that can optimize a set of solutions to produce an archive of diverse, high-performing and reproducible solutions.
Finally, we provide more details regarding the \textit{Reproducibility Improvement Mechanism} on which \name relies.

\subsection{Problem Definition}

As explained in Section~\ref{sec:UQD}, the Uncertain QD problem can be formalized as a multi-objective problem where the descriptor variances are minimized, and the expected fitness inside each cell is maximized (see Problem~\ref{problem:eq:spread}).
However, this optimization problem is computationally challenging to solve directly. 
Instead of minimizing the variance in the descriptor space, we propose to maximize the probability that the descriptor of each solution belongs to its designated cell. 
Thanks to this, problem~\ref{problem:eq:spread} becomes:
\begin{align}
\label{problem:eq:proba}
\begin{split}
    \text{For each cell $\cell_i$, } &\text{find a solution $\params_i$ that:}\\
    &\mbox{maximize}\: \expect{\fitness_i \sim \distFitParamI}{ \fitness_i } \\
    &\mbox{maximize}\: \proba{ \bd_i \in \celltogo } \\
    &\mbox{subject to}\:  \expect{ \bd_i \sim \distDescParamI }{\bd_i}\in\cell_i
\end{split}
\tag{P2}
\end{align}

The descriptor variance, as expressed in Problem~\ref{problem:eq:spread}, is correlated with the new probability objective.
By maximizing $\proba{ \bd_i \in \cell_i }$, we minimize a lower bound on the trace of the covariance matrix (see proof in Appendix~\ref{appendix:proofs}).

\subsection{\name Outline}

In this work, we propose the \textit{Archive Reproducibility Improvement Algorithm} (\name{}), that addresses problem~\ref{problem:eq:proba}.
\name takes as input a set of solutions; this set of solutions can be an archive returned by a QD algorithm, or even a single optimized solution returned by an ES algorithm.
Thus, \name is a plug-and-play approach which can be applied to any set of solutions returned by a QD or ES algorithm; it returns an archive with high-performing and diverse solutions that are also highly reproducible.
The outline of \name can be divided into two successive phases, all detailed in Algorithm~\ref{algo:qdr}:
\begin{enumerate}[leftmargin=*]
    % \item The \textit{Initialization} phase: which provides a collection of solutions to start the reproducibility optimization process.
    \item The \textit{Reproducibility Improvement} phase: improves the reproducibility of the solutions provided as input, by using a \textit{Reproducibility Improvement Mechanism}.
    \item The \textit{Archive Completion} phase: uses the same Reproducibility Improvement Mechanism to find high-performing and reproducible solutions in the archive cells that are still empty.
\end{enumerate}

\subsubsection{Reproducibility Improvement Phase}

The Reproducibility Improvement phase aims at improving the reproducibility of the solutions provided as input to the algorithm.
To that end, for each solution in the input set, we reevaluate it $\numsamples$ times and we calculate their mean descriptor; the mean descriptor is defined as the average of the descriptors observed from $\numsamples$ reevaluations of the same solution.
Then we check which cell $\celltogo$ contains this mean descriptor.
And we use our Reproducibility Improvement Mechanism to optimize for the performance and reproducibility of the solution with respect to the cell $\celltogo$.
This Reproducibility Improvement Mechanism consists of using an ES to optimize this solution for $\numGradientAscent$ steps; more details are provided below.
Those steps are detailed in Algorithm~\ref{algo:qdr} on lines~\ref{algo:qdr:line:reprod:beg}-\ref{algo:qdr:line:reprod:end}.

\subsubsection{Archive Completion Phase}

At the end of the Reproducibility Improvement Phase, there are still some cells in the grid that are not covered.
For example, there is no input solution whose mean descriptor ends in a cell, then this cell is still empty at the end of the previous phase.
Therefore, after having improved the reproducibility of all initial solutions, we have two complementary sets of cells: $\explored$ and $\togo$.
The $\explored$ contains all the cells that have been used as targets by \reprodFunc so far.
The purpose of this phase is to find controllers populating the remaining cells to explore $\togo$.
%
% To do that, we explore the remaining cells from $\togo$ step by step, starting from the solutions in $\explored$.
To do that, we explore the remaining cells from $\togo$ step by step, as depicted on Figure~\ref{fig:teaser:aria}.

At each iteration loop, we start by selecting a pair of cells $\cellstart$, and $\celltogo$ satisfying three conditions: (i) $\cellstart\in\explored$, (ii) $\celltogo\in\togo$, and (iii) $\cellstart$ and $\celltogo$ are adjacent. 
Then, we take the initial solution $\meanoptInit$ from $\cellstart$, and use \reprodFunc to optimize its fitness and its probability to fall into $\celltogo$.
In the end, we collect the solution resulting from this optimization procedure $\meanopt_{\mbox{togo}}$.
%
% Then we run the \reprodFunc function starting from the solution of the explored cell $\cellstart$, and targeting the adjacent cell $\celltogo$, and collect the solution resulting from the optimization procedure $\meanopt_{\mbox{togo}}$.
%
%
% Finally, we add $\meanopt_{\mbox{togo}}$ to the target cell $\celltogo$; and we add $\celltogo$ to the set of explored cells $\explored$, and remove it from $\togo$.
%
Note that nothing prevents the mean descriptor of the optimized solution $\meanopt_{\mbox{togo}}$from not being in the target cell $\celltogo$\footnote{All our metrics and plots take this into account. In other words, if the mean descriptor $\meanopt_{\mbox{togo}}$ is not in $\celltogo$, then none of our metrics and plots will consider it in $\celltogo$.}.
This can happen if the descriptor cell is not reachable, or if the Reproducibility Improvement Mechanism ends too early.
In any case, we add $\meanopt_{\mbox{togo}}$ to the target cell $\celltogo$; and we add $\celltogo$ to the set of explored cells $\explored$, and remove it from $\togo$.
We repeat the previous steps until there are no more cells to explore.
All those details are provided in Algorithm~\ref{algo:qdr} from line~\ref{algo:qdr:line:completion:beg} onward.

% Which lines in algorithm?

\subsection{Reproducibility Improvement Mechanism}

\begin{algorithm}[t]
  \small

  \caption{ \reprodFunc}
  \newcommand\algorithmicitemindent{\hspace*{\algorithmicindent}\hspace*{\algorithmicindent}}

    \label{algo:reprod-improver}

  \begin{algorithmic}[1]
\footnotesize
  \State{\textbf{Input:} Initial genotype $\meanoptInit$, Cell $\celltogo$, Number gradient steps $\numGradientAscent$, Number samples per gradient step $\numsamples$, Standard deviation sampling $\sigma$}

  \State{$\meanopt \gets \meanoptInit$}

  \For{$\mbox{step} = 1 \rightarrow \numGradientAscent$} 
  
      \For{$s = 1\rightarrow \numsamples$}
        % \State{Sample $\noise_s \sim \cN(\vec 0, \iI)$}
        \State{Sample $\noise_s \sim \cN(\vec 0, \iI)$}
        % \State{$\fitness_s, \bd_s \gets$ stochastic evaluation of genotype $\params_s = \meanopt + \sigma \noise_s$}
        \State{$\fitness_{s}, \bd_{s} \gets$ stochastic evaluation of genotype $\params_{s} = \meanopt + \sigma \noise_{s}$}
      \EndFor
    
      % \State{$\left( \noiseSort_s \right) \gets$ sort all $\left( \noise_s \right)$ by ranking the $(\fitness_s, \bd_s)$ using order "$\preceq$"}
    
      \State{$\left( \utility_{s} \right) \gets$ rank-based utility values depending on their order for objective $\obj_{\celltogo}$ (Eq.~\ref{eq:objective:maximize})}
    
      \State{$\gradientEstimate \gets \frac{1}{\sigma} \sum_{s=1}^{\numsamples} \utility_s \noise_s $}

      \State{$\meanopt \gets$ perform gradient ascent step on $\meanopt$ using $\gradientEstimate$}
    
    % \item[]
  \EndFor
      \State{\Return final genotype $\meanopt$}
  \end{algorithmic}
\end{algorithm}

As explained in the previous section, the Reproducibility Improvement Mechanism takes as input a solution to be optimized, and a target cell $\celltogo$.
This mechanism aims at optimizing the solution's expected fitness and probability of belonging to $\celltogo$, while also satisfying the constraint: having its mean descriptor belonging to $\celltogo$ (see Problem~\ref{problem:eq:proba}).
To address this constrained optimization problem for a particular target cell $\cell=\celltogo$ and its associated centroid $\centroid$, we rely on the ES of~\citet{ESScalableAlternative}, detailed in Section~\ref{sec:nes} and in Algorithm~\ref{algo:reprod-improver}.
%

% Our \reprodFunc procedure takes as input an initial solution $\meanoptInit$ and a target cell $\celltogo$.
%
% It aims at optimizing the solution's expected fitness, and probability of belonging to the cell (see Problem~\ref{problem:eq:proba}), while also satisfying the constraint: having its mean descriptor belonging to the cell.

% \subsubsection{Objective Function and Solutions Ranking}

\label{sec:objectivefunction}

\newcommand{\pair}{\vec p}

This ES algorithm iteratively updates a solution $\meanopt$ by (1) sampling and evaluating neighboring solutions in the search space following a Gaussian distribution, (2) ranking those solutions by order of preference, and (3) updating $\meanopt$ following the estimated rank-based gradient (see Section~\ref{sec:nes}).
%
% The ranking of the solutions is performed following those three rules:
In order to improve the reproducibility of $\meanopt$, we rank the solutions in the following manner:
\begin{itemize}[leftmargin=*]
    \item All solutions whose evaluated descriptor is in the cell, are ranked higher than those outside of the cell.
    \item Among the solutions whose descriptors are outside the cell, solutions are ranked according to their distance to the center of the cell: the closer the better.
    \item Among the solutions whose descriptors are inside the cell, solutions are ranked according to their fitness: the higher the better.
\end{itemize}

More formally, this is equivalent to maximizing the following objective function, which satisfies all the above ranking rules:\footnote{For this formulation, we assume that $\fitness$ is bounded as it simplifies notations. However, this assumption is not very strong, because if the fitness is unbounded, then $\fitness_{\mbox{min}}$ can be set to the minimal fitness encountered so far.}
\begin{equation}
\begin{aligned}[b]
    \noisyobj_{\cell} (\params) 
    &=  \obj_{\cell}(\fitness, \bd) \quad \text{with $\fitness$ and $\bd$ sampled from $\distFitParam,\distDescParam$} \\
    &=  \begin{cases}
    \fitness \quad &\mbox{if } \bd\in\cell \\
    \fitness_{\mbox{min}} -\normeucl{\bd - \centroid} \quad &\mbox{if } \bd\notin\cell
    \end{cases} \label{eq:objective:maximize}
\end{aligned}
\end{equation} 
% \footnotetext{For this formulation, we assume that $\fitness$ is bounded as it simplifies notations. However, this assumption is not very strong, because if the fitness is unbounded, then $\fitness_{\mbox{min}}$ can be set to the minimal fitness encountered so far.}
% \footnotetext{if the fitness is unbounded, $\fitness_{\mbox{min}}$ can simply correspond to the minimal fitness encountered by the optimization process so far, and can be updated at each iteration if new lower values are encountered. In any case, we use a rank-based fitness shaping afterwards; so only their rank matters. }
%
% Note that the function $\noisyobj(\cdot)$ is stochastic, while the function $\obj(\cdot)$ is a deterministic function of the sampled fitness and descriptor.
%
% Indeed, by ranking solutions with this objective function, we abide by all the rules expressed above.
%
This structure of objective function has been proposed in the evolutionary literature as a way to address constrained optimization problems~\cite{coello2002theoreticalconstrainttechniques,hansen2016cmaestutorial}.
Note that the function $\noisyobj(\cdot)$ is stochastic, while the function $\obj(\cdot)$ is a deterministic function of the sampled fitness and descriptor.
Additionally, if we make the assumption that the fitness is bounded, it is provable that by maximizing the expectation of $\noisyobj_{\celltogo}(\cdot)$, we maximize a lower bound on the two objectives of Problem~\ref{problem:eq:proba}, namely: the expected fitness and the probability of belonging to a descriptor cell (see proof in Appendix~\ref{appendix:proofs}).

\section{Experimental Setup}

\subsection{Metrics}

\label{sec:metricsol}

\subsubsection{Expected Fitness (\ef) and (Corrected) QD-Score}

\label{sec:metricssol:ef}

The Expected Fitness (\ef)~\cite{flageat2022benchmarking} characterizes the average performance of a solution with respect to the task.
As given in Problems \ref{problem:eq:spread} and~\ref{problem:eq:proba}, it is one of the quantities explicitly maximized for in our algorithms.
To estimate the expected fitness of one solution, we reevaluate it $\numReplications$ times, and average the obtained fitnesses.
\begin{align}
\label{eq:expectedfitness}
    \mbox{\ef}(\params) = \expect{\fitness \sim \distFitParam}{\fitness} \approx \frac{1}{\numReplications} \sum_{m=1}^{\numReplications} \fitness_m \quad \mbox{with } \fitness_m \sim \distFitParam
\end{align}

The (Corrected) QD-Score~\cite{Pugh2015, flageat2022benchmarking} characterizes both the quality and diversity of an archive.
It corresponds to the sum of all the expected fitnesses of solutions contained in an archive.
To cope with negative values, all expected fitnesses are first normalized between $0$ and $1$ before being summed.

\subsubsection{Negated Descriptor Variance (\nvs) and Variance Score (V-Score)}
\label{sec:metricssol:nvs}

The \nvs~\cite{flageat2023uncertain} characterizes the spread of the descriptor samples.
Mathematically, it corresponds to the negated trace of the covariance matrix of the descriptor samples.
If a solution has a high \nvs score, i.e. an \nvs score close to $0$, it means that its sample descriptors are (in average) concentrated around their mean.  
This metric comes from the variance objective that we minimize in our original Uncertain QD problem (see Problem~\ref{problem:eq:spread}).
We evaluate it with the following unbiased estimator (where $\bar{\bd} = \frac{1}{\numReplications} \sum_{m=1}^{\numReplications} \bd_m$):
\begin{align*}
    \mbox{\nvs}(\params) &= -\sum_{j} \var{ \bd \sim \distDescParam }{ \bd^{(j)} } \\
    &\approx \frac{-1}{\numReplications - 1} \sum_{m=1}^{\numReplications} \normeucl{ \bd_m - \bar{\bd} }^2 \quad \text{with } \bd_m\sim\distDescParam
\end{align*}
%
% This estimator follows from Lemma~\ref{lemma:varsum}.

The Variance Score (V-Score) is the \nvs equivalent of the QD-Score.
Instead of considering normalized expected fitnesses, the V-Score corresponds to the sum of normalized \nvs scores of each cell in the archive.

\subsubsection{Probability of Belonging to the Mean Descriptor Cell and Probability Score (P-Score)}

\label{sec:metricsol:proba}

% The probability of belonging to a cell defines the proportion of a solution's descriptors that are in the same cell as the solution.
The probability of belonging to a cell is defined as: the probability that the descriptor of a solution belongs to its designated cell.
This metric characterizes a solution's reproducibility; it corresponds to the probability objective that our algorithm maximizes (see Problem~\ref{problem:eq:proba}).
Given a solution $\params$, we first determine which cell $\cell_i$ the mean descriptor $\bar{\bd}$ belongs to; we then estimate $\proba{ \bd \in \cell_i }$ by reevaluating its descriptor $\numReplications$ times and calculating the proportion of samples ending up in the cell:
\begin{align*}
    &\proba{ \bd \in \cell_i } \approx \frac{1}{\numReplications} \sum_{m=1}^{\numReplications} \iverson{ \bd_m \in \cell_i } \quad \mbox{with } \bd_m \sim \distDesc \mbox{ and } \bar{\bd}\in\cell_i \\
&\text{where $\iverson{\cdot}$ works as an Iverson bracket: } \iverson{P} = \begin{cases} 1 \text{ if $P$ is true} \\ 0 \text{ if $P$ is false} \end{cases}
\end{align*}

Similar to the QD-Score and V-Score, we consider the Probability Score (P-Score).
This P-Score equals the sum of the probability value obtained from each solution in the final archive.

\subsubsection{Archive Profile}

\begin{figure}[t]
    \centering
    \includegraphics[width=0.35\textwidth]{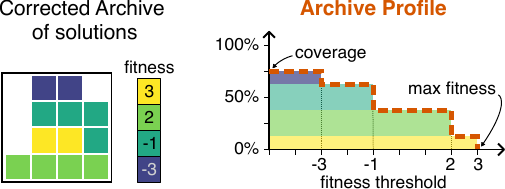}
    % \vspace{-4mm}
    \caption{Illustration of the derivation of the Archive Profile (dashed orange line on the right) from a Corrected Archive (left). We also indicate how to retrieve the archive coverage and max fitness from the Archive Profile graph.}
    \label{fig:archive_profile_illustration}
\end{figure}

\newcommand{\fthreshold}{f_{\mbox{thres}}}

% The purpose of our approach is to optimize the quality and diversity of the final archive returned by the algorithm.
% %
The \textit{Archive Profile} is a holistic way of visualizing the fitness distribution of the resulting archive~\cite{fontaine2021differentiable, flageat2022benchmarking}.
More precisely, for a given fitness threshold $\fthreshold$, the Archive Profile corresponds to the number of archive cells $\cell$ with a solution having a better expected fitness.
In our experiments, we normalize it, by dividing the above quantity with the total number of cells.
This way, we estimate the proportion of grid cells filled with solutions having a fitness greater than $\fthreshold$:
\begin{align*}
\Call{ArchiveProfile}{}(\fthreshold) = \frac{1}{\numCells}\sum_{\mbox{cell}\; \cell} \iverson{ \fitness_{\cell} \geq \fthreshold }
\end{align*}
where $\iverson{\cdot}$ is the Iverson bracket, as described in section~\ref{sec:metricsol:proba}.

The Archive Profile exhibits several properties of interest for the archive.
For instance its leftmost value on the $y$ axis corresponds to the archive's total coverage; while the intersection of the curve with the $x$ axis indicates the highest fitness found in the archive (see Fig.~\ref{fig:archive_profile_illustration} for an illustration).
%
% Moreover, when the considered $m$ is the expected fitness, then the area under the curve is proportional to the QD-Score~\citep{Benchmark paper}.

\subsection{Corrected Archives}

\label{sec:archivemetrics}

\newcommand{\metric}{m}
\newcommand{\archMetric}[1]{\psi_{\geq} \left( { #1 } \right) }
The archives returned by QD algorithms may contain degenerate solutions, whose descriptor may significantly differ from the expected descriptor.
Consequently, in such archives, the cell a solution is placed in may not be the cell of its expected descriptor; we designate those archives as: \textit{Illusory Archives}.
To alleviate this issue, our results are only presented in terms of \textit{Corrected Archive}~\cite{justesen2019adaptivesampling, flageat2020deepgrid, flageat2022benchmarking}.
Given an archive or set of genotypes returned by a QD algorithm, all genotypes are reevaluated $\numReplications$ times, and placed in a new archive, called "Corrected Archive" based on their mean descriptor $\meanbd = \sum_{s=1}^{\numReplications} \bd_s$; and in each cell, we only keep the solution having the best estimated expected fitness (see Formula~\ref{eq:expectedfitness}).
The main difference between an illusory archive and its corrected archive is depicted on Fig.~\ref{fig:aria-improvement-over-lucky}.

Note that in this work, all the archive metrics detailed above are "corrected"~\cite{flageat2022benchmarking}: they are only applied on the corrected archive.
Consequently, we omit the prefix "Corrected" when talking about the QD-Score, the NVS-Score and the P-Score.

% \subsubsection{QD-Score}

% The QD-Score~\cite{Pugh} characterizes both the quality and diversity of an archive.
% %
% It corresponds to the sum of all the expected fitnesses (see section~\ref{sec:metricssol:ef}) of solutions contained in an archive.
% %
% To cope with negative values, all expected fitnesses are first normalized between $0$ and $1$ before being summed.

% \subsubsection{Variance Score (V-Score)}

% The Variance Score (V-Score) is the \nvs equivalent of the QD-Score.
% %
% Instead of considering normalized expected fitnesses, the V-Score equals the sum of normalized \nvs scores (see section~\ref{sec:metricssol:nvs}) of each cell in the archive.

% \subsubsection{Probability Score (P-Score)}

% Similarly to the QD-Score and V-Score, we consider the Probability Score (P-Score).
% %
% This P-Score equals the sum of the probability value obtained from each solution (as calculated in section~\ref{sec:metricsol:proba}).

\subsection{\name Variants}

\begin{table}[t]
    \centering
    \small

    \begin{threeparttable}
    \begin{tabular}{l c l l}
    \toprule

% Variant & \begin{tabular}{@{}l@{}}
%                     \\
%                    Task Solver \\
%                  \end{tabular}
%                  & Has Fitness \\

Baseline & Reprod\tnote{$\ast$} & Fitness(es) & Descriptor \\
        \midrule
    	\addlinespace

\momer & \cmark & $\begin{cases} \bar{\fitness} = \frac{1}{\numreeval} \sum_{s=1}^{\numreeval} \fitness_{s} \\ -\mbox{\nvs} \end{cases}$ & $\bar{\bd} = \frac{1}{\numreeval} \sum_{s=1}^{\numreeval} \bd_{s}$ \\
\mesar & \cmark & $\bar{\fitness} - \nvs$\tnote{$\ast\ast$} & $\bar{\bd}$ \\

\addlinespace

\mesa & \xmark & $\bar{\fitness}$ & $\bar{\bd}$ \\
\mapelites & \xmark & $\fitness$ & $\bd$ \\

    \bottomrule
    	
    \end{tabular}
    
    \begin{tablenotes}
         \item[$\ast$] Does the baseline optimize also for reproducibility?
         \item[$\ast\ast$] To get values of the same magnitude, the two terms are first normalized before being summed.
     \end{tablenotes}
    \end{threeparttable}

\caption[Summary]{Characteristics of the baselines under study.}
    \label{table:variants2}

\end{table}

\name can take as input any set of genotypes.
As explained above, in \name, this set of genotypes goes through a Reproducibility Improvement phase, and is then used as a basis for finding new reproducible solutions in the rest of the archive.
To study the effect of the initial set of genotypes, we initialize \name with different kinds of input.
The corresponding variants are the following:
\begin{itemize}[leftmargin=*]
    \item \textbf{\nameME} which takes as input an archive returned by \mapelites{}.
    This way we initialize the \name{} algorithm with high-performing controllers that cover different areas of the descriptor space.
    %
    % The potential issue behind this method, is that \mapelites{} does not present any mechanism to handle the stochasticity of descriptors and fitnesses; which means that many cells may be populated with non-reproducible solutions.
    %
    % This problem is tackled by the next phase of \name{}: \textit{reproducibility improvement}.

    \item \textbf{\namePGA} which takes as input an archive returned by \pgame~\cite{nilsson2021pgamapelites}.
    While this kind of environment is only compatible with QD-RL tasks~\cite{tjanakabryon2022approximating}, it is expected to return better archives when the solutions parameterize high-dimensional deep-neural network controllers.
    
    \item \textbf{\nameES} runs an Evolution Strategy algorithm to maximize the performance of a single solution.
    %
    % In this process, the objective we maximize is the expected fitness only, which differs from the objective introduced before (see Eq.~\ref{??}).
    The maximized objective is the expected fitness only; and we use the same optimization procedure as before (see section~\ref{sec:nes}).
    %
    % The advantage of this method compared to a plain \mapelites{}, is that it presents mechanisms handling the stochasticity of the problem.
    %
    % In addition, it has been proven effective to train high-dimensional controllers~\cite{ES Scalable alternative}.
    
    % Which one?
    % How do we use it?
\end{itemize}

Also, we introduce a new variant, called "Linear Reproducibility Improvement Mechanism" (\textbf{\linearrim}), which does not optimize for the same objective function as \name.
With this variant, we intend to show that the constraint-based objective function used by the Reproducibility Improvement Mechanism of \name is also of importance.
In particular, the objective function used by \linearrim is equal to the sum of the two (normalized) components of the objective function present in equation~\ref{eq:objective:maximize}.
The rest of the optimization procedure remains the same; and, similarly to \nameME, it takes as input an archive generated by \mapelites.

% Who introduced the Corrected Archive?

\subsection{Baselines}  \label{sec:baselines}

The first baseline we consider for comparison with \name is the \textbf{\mapelites} algorithm, which samples only one descriptor and fitness per solution evaluation.
This lower baseline does not take into account the uncertainty aspect of the problem.
As a result, the illusory archive returned by \mapelites and its corresponding corrected archive may substantially differ (see Fig.~\ref{fig:aria-improvement-over-lucky}).
% With this baseline, we intend to prove the benefit of using \name with a \mapelites initialization, over using just a \mapelites.

We intend to also compare our variants of \name to several algorithms designed for Uncertain QD problems; these algorithms are detailed below, and their characteristics are summarized in Table~\ref{table:variants2}.
\begin{itemize}[leftmargin=*]
    \item \mapelites Sampling (\textbf{ME-Sa})~\cite{justesen2019adaptivesampling} works as \mapelites, except that for each solutions, it samples several fitnesses and descriptors, and use them to calculate the mean fitness and mean descriptor.
    This way, \mesa gets better estimates of a solution's expected fitness and descriptor; and the archive returned by \mesa is expected to be more similar to its corrected archive.
    
    \item \mapelites Sampling with Reproducibility  consideration (\textbf{\mesar}) is similar to \mesa, except that the fitness of each solution is modified to take into account the reproducibility problem.
    The fitness of \mesar adds a term that penalizes solutions having a high \nvs score (see Table~\ref{table:variants2}) on top of the expected task fitness.
    
    \item Multi-Objective \mapelites with Reproducibility consideration (\textbf{\momer}).
    As the problem to solve is initially a multi-objective problem (see Problem~\ref{problem:eq:spread}), we propose to also study the results obtained by a variant of Multi-Objective \mapelites (\mome)~\cite{pierrot2022mome}.
    The two fitnesses used by \momer are the two terms of the objective used by the \mesar baseline.
    We choose a Pareto front length of $50$, as in the introductory work of \mome.
    In this case, the \momer algorithm is not directly comparable to our approach, as the archive it returns may contain way more solutions than all other algorithms.
    Indeed, up to $50$ solutions may be stored in each cell of the \momer archive.
    To make this approach comparable, at analysis time, in each cell, we only pick the solution maximizing the \mesar objective (see Table~\ref{table:variants2}).
\end{itemize}

\subsection{Tasks}

% \begin{figure}
%     \centering
%     \includegraphics[width=3cm]{example-image-a}
%     \caption{Illustration of the three tasks under study. From left to right: Arm, Ant, and Walker environments.}
%     \label{fig:tasks}
% \end{figure}

We compare the performance of all algorithms on three tasks, commonly used in Uncertain QD Settings \cite{flageat2023uncertain}: the noisy 8 Degrees of Freedom (DoF) Arm, as well as the Ant Omni-directional and Walker Uni-directional Tasks.
%
% Those three tasks are illustrated in Figure~\ref{fig:tasks}
%
% Those task details are summarized in Table~\ref{table:hc_qd_tasks}.

\subsubsection{Noisy 8-DoF Arm}

\newcommand{\variance}[1]{\mbox{Var} \left( {#1} \right)}

The Arm task consists of a planar robotic Arm, that we control to reach different final positions \cite{Cully2018}.
The solution consists of the angular positions of the successive actuators $\params = \left( \theta_i \right)_{i\in\left[1..8\right]}$.
The fitness promotes solutions having an homogeneous set of angles, and some Gaussian noise is added to it to make its evaluations uncertain.
Also, the descriptor corresponds to the final position $(x_E, y_E)$ of the end-effector perturbed by some Gaussian noise.
The fitness and descriptor can be expressed as follows, with $\sigma_{\fitness}=\sigma_{d}=0.01$:
    \begin{align}
    \label{eq:armfitdesc}
        \fitness(\params) = -\variance{\theta_i} + \mathcal{N}\left( 0, \sigma_{\fitness}^2 \right) \quad \bd(\params) = \begin{pmatrix}
    x_E \\ y_E
    \end{pmatrix} + \mathcal{N} \left( \vec 0, \sigma_{d}^2 \mat I \right)
    \end{align}
% %
% If we write $(x_E, y_E)$ the position of the end-effector, then the descriptor returned by the algorithm is:
% %
% \begin{align}
%     \bd(\params) = \begin{pmatrix}
%     x_E \\ y_E
%     \end{pmatrix} + \mathcal{N} \left( \vec 0, \sigma_{d}^2 \mat I \right)
% \end{align}

\subsubsection{Ant Omni-directional}

The Ant Omni-directional task consists of finding high-dimensional controllers bringing a four-legged robot to diverse final positions.
The task fits in the QD-RL setting~\cite{tjanakabryon2022approximating, flageat2022benchmarking}; it can be modeled as a Markov Decision Process (MDP), and the task fitness equals the non-discounted sum of rewards over an entire episode.
In the case of the Ant Omni task, the reward promotes solutions who don't die prematurely, while it penalizes high energy consumption $\fitness = \sum_{t=1}^{T} r_{survive} - r_{energy}$ with $T=100$.
Also, the descriptor corresponds to the final position of the robot: $\bd=(x_T, y_T)$.
We use the Ant Omni task as implemented in the \qdax library~\cite{lim2022accelerated}, which itself relies on the Ant environment from \brax~\cite{brax2021github}.
The learned controllers are closed-loop deep neural network policies, with two hidden layers of size 64, which output the torques to apply on the robot joints at each timestep.
In this task, the noise is applied on the initial joint positions and velocities.

\subsubsection{Walker Uni-directional}

The Walker Uni-directional task consists of finding diverse ways to move forward as fast as possible.
Similarly to the Ant Omni task, it fits in the QD-RL setting, and we also rely on the implementation from \qdax using \brax.
Furthermore, the noise and controllers considered are the same as for the Ant task.
Nonetheless, the fitness and descriptor differ from the Ant task.
Indeed, here the reward presents an additional term promoting forward displacement: $\fitness = \sum_{t=1}^{T} r_{forward} + r_{survive} - r_{energy}$ with $T=1000$.
The descriptor here characterizes the proportion of time each leg touches the ground: $\bd = \frac{1}{T}\sum_{t=1}^{T} \left( C_1 (t), C_2 (t) \right)$, where $C_i(t)$ equals $1$ if leg $i$ touches the ground at timestep $t$, and equals $0$ otherwise.

\subsection{Implementation Details}

For each task under study, the descriptor space is two-dimensional, and is discretized in a $32\times 32$ grid.
The archives given as input to \nameME and \namePGA are generated with respectively \num[round-precision=1,round-mode=figures,
     scientific-notation=true]{2000000} and \num[round-precision=1,round-mode=figures,
     scientific-notation=true]{500000} evaluations.
% The \textsc{ME} and \textsc{PGA} variant of \name are respectively granted a maximal budget of $2\times 10^6$ and $5\times 10^5$ evaluations for their initialization phase.
%
At each iteration of the reproducibility improvement function, $\numsamples=2{,}048$ samples are generated for the Arm and Ant Omni tasks; and $\numsamples=1{,}440$ is chosen for the Walker task as it tends to be computationally heavier.
We also use mirror sampling, so the Reproducibility Improvement Mechanism performs respectively $4{,}096$ and $2{,}880$ evaluations per iteration.
Moreover, the samples are drawn with a standard deviation of $\sigma=0.005$, $0.02$ and $0.01$ respectively for the Arm, Ant and Walker tasks.
Finally, the number of gradient steps taken by the reproducibility improvement mechanism is $\numGradientAscent=100$ for the Arm and Ant tasks; and $\numGradientAscent=75$ for the Walker task.

The \momer, \mesar and \mesa variants use a batch size of 128, with each solution reevaluated $\numreeval=32$ times, which makes a total sampling size~\cite{flageat2023uncertain} of $4{,}096$ per iteration.
Hence, we also use this value as batch size for the \mapelites baseline.
All those baselines have a total budget equal to the maximal number of evaluations done by \name variants, i.e. \num[round-precision=2,round-mode=figures,
     scientific-notation=true]{428228608} for the Arm and Ant tasks, and \num[round-precision=2,round-mode=figures,
     scientific-notation=true]{229386240} for the Walker task.

At analysis time, each solution is reevaluated $\numReplications=1{,}024$ times to compute the archive profile.
This value is used for all metrics.

All our implementation is based on the \qdax framework~\cite{lim2022accelerated}, except for the reproducibility improvement mechanism, which is adapted from the evolution strategy implementation from \brax~\cite{brax2021github}.
Each experiment was run on an NVIDIA RTX A6000; with this hardware, \name took respectively 20 minutes, 10 hours, and 18.5 hours to run the Arm, Ant and Walker tasks.
We run each variant for 10 replications; and we evaluate the statistical significance of our comparisons using the Wilcoxon rank-sum test, with a Holm-Bonferroni correction \cite{holm1979simple}.
To facilitate the replications of the results, we stored our code and dependencies in a singularity container \cite{kurtzer2017singularity}, and made it available with the code at: \url{https://bit.ly/aria-gecco}.

% \begin{itemize}
%     % \item Adam Optimizer
%     % \item parameters of PGA? 
%     \item all the other algorithms are in appendix.
% \end{itemize}

\section{Results}

\begin{figure}[t]
    \centering
    \includegraphics[width=0.49\textwidth]{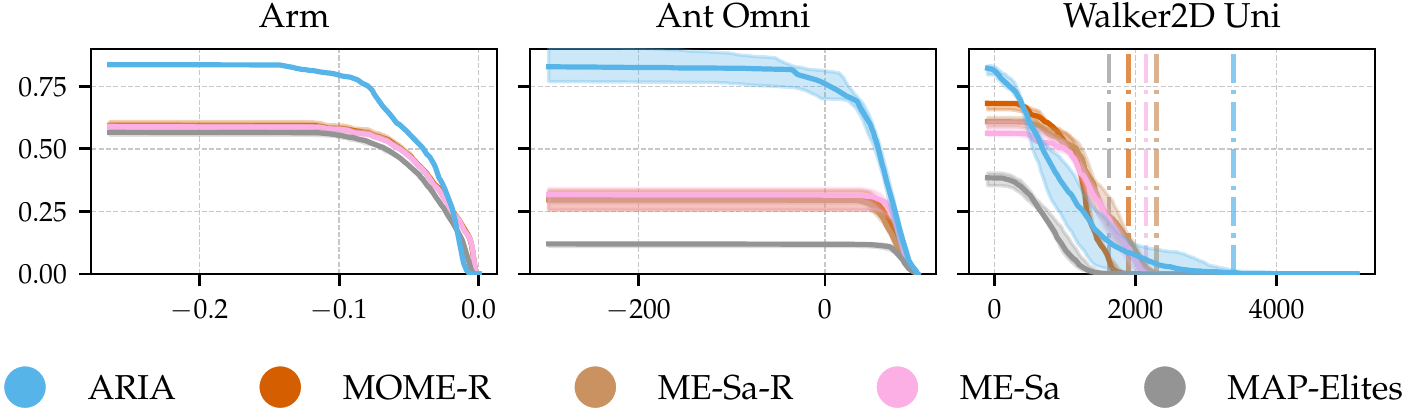}
    \vspace{-7mm}
    \caption{Archive Profiles of \name and other baselines. 
    For clarity purposes, only one variant of \name is shown per plot.
    From left to right, the plotted variants of \name are respectively: \meinit, \meinit and \pgainit.
    The bold line represents the median, and the shaded area corresponds to the inter-quartile range.
    The vertical lines on the Walker task represent the median max fitness.
    }
    \label{fig:01_archive_profiles}
\end{figure}

\begin{table*}[t]
\small
    \centering

    \begin{tabular}{lrrrrrrrrr}
\toprule
       \multirow{2}{*}{Algorithm} & \multicolumn{3}{c}{Arm task} & \multicolumn{3}{c}{Ant Omni-directional task} & \multicolumn{3}{c}{Walker Uni-directional task} \\ \cmidrule(lr){2-4} \cmidrule(lr){5-7} \cmidrule(lr){8-10} & QD-Score & V-Score & P-Score & QD-Score & V-Score & P-Score & QD-Score & V-Score & P-Score  \\
\midrule

  \textsc{ARIA - ME init} & {\bfseries 722.41} &               426.83 & {\bfseries 653.37} & {\bfseries 765.69} &   {\bfseries 818.14} & {\bfseries 453.94} &              81.04 &                  776.81 &               561.96 \\
\textsc{ARIA - PGA init} &             \xmark &               \xmark &             \xmark &             679.06 &               756.64 &             414.12 &             180.72 &      {\bfseries 839.87} &   {\bfseries 634.06} \\
 \textsc{ARIA - ES init} &             587.30 &   {\bfseries 428.22} &             647.37 &             581.22 &               613.61 &             279.80 & {\bfseries 248.14} &                  701.43 &               466.58 \\

\addlinespace
 
     \textsc{Linear RIM} &             704.68 &               394.23 &             542.58 &             610.28 &               619.30 &             341.06 &             115.13 &                  642.80 &               373.22 \\

     \addlinespace

         \textsc{MOME-R} &             538.62 &               303.57 &             258.03 &             287.90 &               268.79 &              47.77 &             175.75 &                  692.74 &               284.29 \\
        \textsc{ME-Sa-R} &             538.19 &               304.37 &             258.94 &             283.39 &               279.24 &              61.34 &             185.68 &                  613.97 &               174.67 \\
          \textsc{ME-Sa} &             534.06 &               301.37 &             253.61 &             307.26 &               251.20 &              33.40 &             168.70 &                  554.26 &               123.61 \\
     \textsc{MAP-Elites} &             506.34 &               288.63 &             318.59 &             117.91 &                93.65 &              13.44 &              66.87 &                  357.52 &                19.75 \\

\bottomrule
\end{tabular}

    \caption{Median (Corrected) QD-Scores, Variance-Scores and Probability-Scores achieved by each algorithm on the three tasks under study.
    Note that the Arm task is not a QD-RL task~\cite{tjanakabryon2022approximating}, so \pgame cannot be directly applied on it.
    }
    \label{table:results}

\end{table*}

\begin{figure*}[t]
    \centering
    \includegraphics[trim={0 0 0 4mm},clip,width=0.81\textwidth]{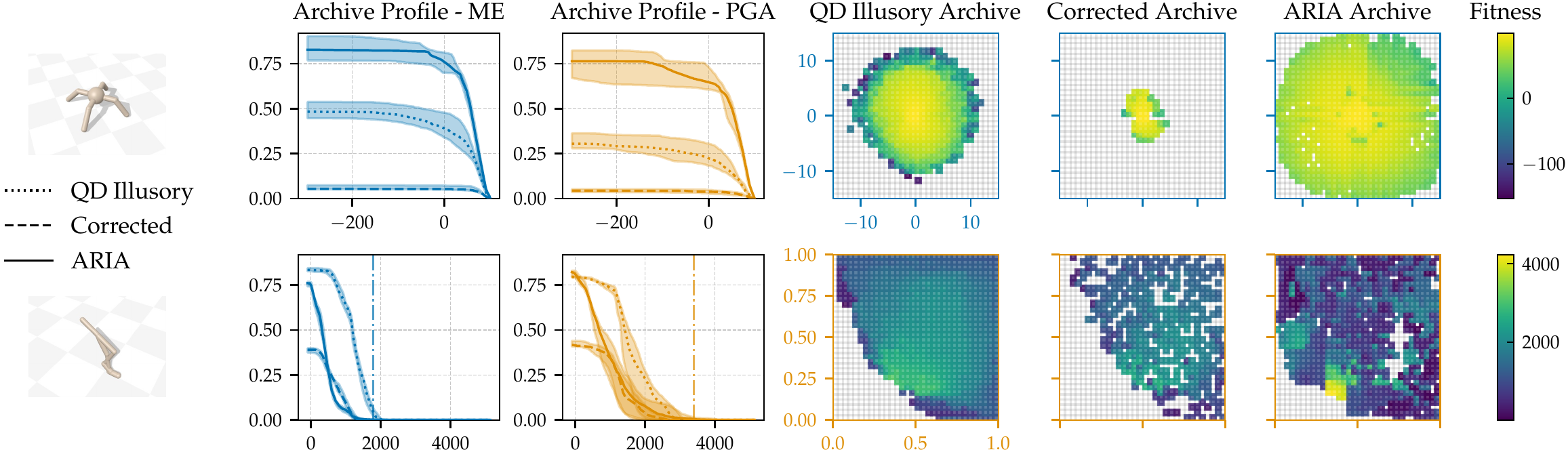}
    \vspace{-3mm}
    \caption{(columns 1 \& 2) Archive Profiles of \nameME and \namePGA on the Ant and Walker tasks; (columns 3 to 5, from left to right) Illusory archive returned by the QD algorithm, its corresponding corrected archive, and the archive obtained after optimization from \name (the chosen archive is the one achieving the median QD-Score out of 10 replications).
    On the walker task, the vertical lines represent the median max fitness achieved by the archive returned by \name.}
    \label{fig:p_01_e_02_improvement_ant_walker}
\end{figure*}

\begin{figure*}[t]
    \centering
    \includegraphics[width=0.85\textwidth]{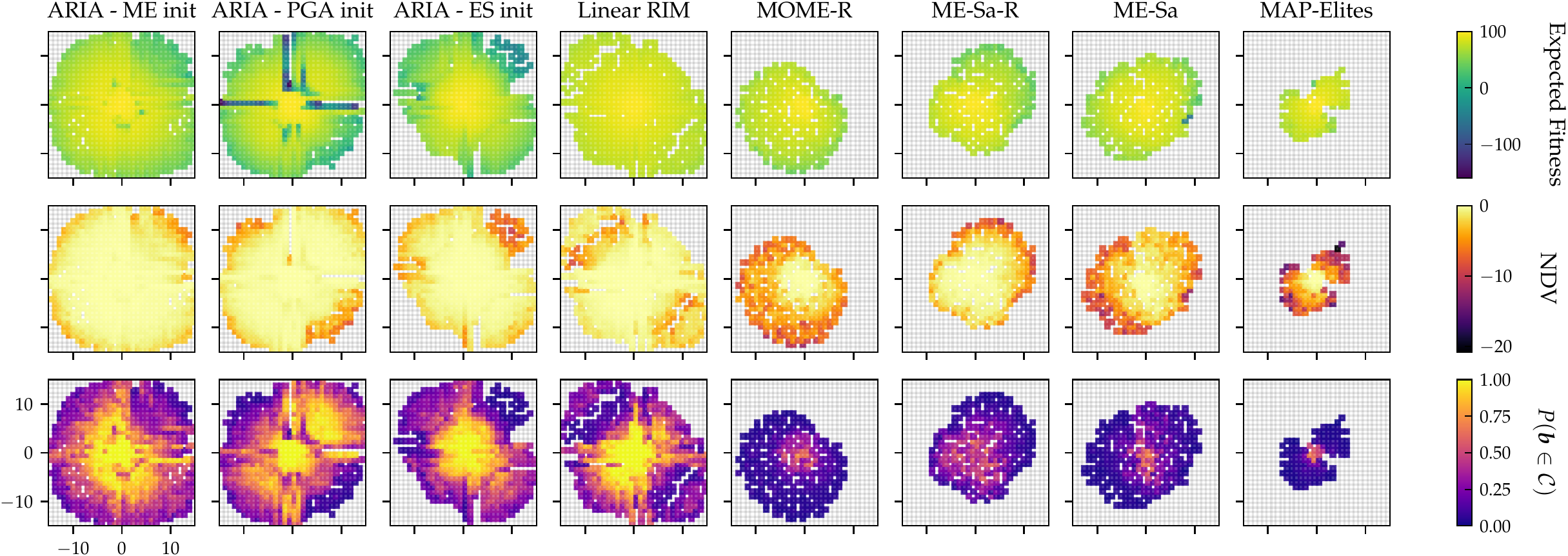}
    \vspace{-3mm}
    \caption{Distribution of the expected fitness, \nvs score, and probability of belonging to cell in an archive returned by each algorithm, on the Ant Task.
The chosen archive is the one achieving the median QD-Score out of 10 replications.}
    \label{fig:repertoires_ant}
\end{figure*}

% The analysis can be divided into two parts.
% %
% On the one hand, we show how \name manages to improve the quality and coverage of an archive returned by a classical QD algorithm.
% %
% Then we show to which extent our approach improves upon other Uncertain QD algorithms, as well as other algorithms improving explicitly for solution's reproducibility (detailed in Section~\ref{sec:baselines}).

% \subsection{}

On the three tasks under study, \name improves substantially the archive profile of the corrected archive given by the initial QD algorithm.
In particular, \name increases the coverage of the corrected archives by at least 50\% on all tasks ($p<2\times 10^{-4}$, Figs.~\ref{fig:aria-improvement-over-lucky} and~\ref{fig:p_01_e_02_improvement_ant_walker}).
Note that we do not intend to reach the results of the illusory archive, as this kind of archive contains many solutions whose evaluated fitness and descriptors differ substantially from their expected fitnesses and descriptors; hence nothing tells if such kind of archive is possible to obtain.
Interestingly, on the Ant task, the coverage obtained by \name is even better than the coverage of the illusory archive ($p<2\times 10^{-4}$).
It is also noticeable that on the Walker task, the corrected archive of \nameMEshort presents more individuals with a fitness higher than 900 compared to \nameME; this phenomenon may be due to the fact that \name optimizes for reproducibility before optimizing for performance, which is corroborated by its high Variance Score in Table~\ref{table:results} and by the extended analysis provided in Appendix~\ref{appendix:extendedanalysiswalker}.
Overall, these results demonstrate that \name improves not only the reproducibility of solution contained in any QD archive, but also can improve their content and sometimes their coverage. 
Videos of our results are available at: \url{https://sites.google.com/view/gecco2023-aria/}.

%%%%%%%%%%%%%%%%%%%%

\nameME and \namePGA achieve a better reproducibility, i.e. higher  V-Scores and P-Scores, than the other baselines which do not use a Reproducibility Improvement Mechanism (Table~\ref{table:results} and Fig.~\ref{fig:repertoires_ant}, $p<0.03$).
This can be explained by the objective function that is explicitly maximized by \name, which puts the focus on the reproducibility before optimizing for the performance.
Indeed, the objective introduced in section~\ref{sec:objectivefunction} prioritizes solutions that are in the cell compared to those that are not.
Furthermore, the ES of the reproducibility improvement mechanism can scale to high-dimensional deep neural networks, making it easier for \name to explicitly optimize for reproducibility.

%%%%%%%%%%%%%%%%%%%%

If we compare the results obtained by the various \nameME, we notice that the performance of each variant is correlated with its initialization.
For example, on the ant task, the poor performance of the \pga initialization archive leads to low-quality results (Fig.~\ref{fig:p_01_e_02_improvement_ant_walker}).
On the contrary, the \meinit variant, which relies on a better corrected archive, manages to reach higher QD-Scores and archive coverage ($p<0.03$).
Surprisingly, the \esinit variant of \name achieves a better QD-score than its \pga counterpart (Table~\ref{table:results}).
However, on that task, the \pga initialization still manages to provide more reproducible archives, which is likely due to its internal optimization process as demonstrated in the literature~\cite{nilsson2021pgamapelites, flageat2022empiricalPGA}.

% \subsection{Comparing \name to Uncertain QD Baselines}

%%%%%%%%%%%%%%%%%%%%

In all tasks, the best QD-Score is obtained by a variant of \name (see Table~\ref{table:results}), and their archive profiles exhibit a higher coverage (Fig.~\ref{fig:01_archive_profiles}).
Also, in the Walker task, the archive profile exhibits a higher maximal fitness.
The worse QD Score obtained for some variants of \name (e.g. \esinit on the Arm task and \meinit on the Walker task) may be due to their initialization.
Indeed, exploring from a single high-performing individual, as done by \nameES, may not be sufficient to find all the elites in the search space of the Arm task; and the high-dimensionality of the solutions in the Walker task prevents the \textsc{ME} initialization from finding high-performing individuals, which appears detrimental in the Walker task.
%

% \subsection{Effect of the Optimization Function}

The constraint-based objective used by the Reproducibility Improvement mechanism is also of importance.
Indeed, \linearrim never manages to outperform \nameMEshort, both in terms of performance and reproducibility.
On the Arm and Ant tasks, it achieves lower QD, Variance and Probability scores compared to \nameMEshort (Table~\ref{table:results}, $p<8\times 10^{-3}$).
On the Walker task, \linearrim reaches a higher QD score compared to \nameME, but its Variance and Probability scores remain significantly lower ($p<2\times 10^{-4}$).

% \subsection{\name Manages to Find good Reevaluated Archives}

% What if we change the objective of the reproducibility improver?

% What if we just run MAP-Elites and try to rereach the lucky cells (in other words, we remove the line of reevaluation and mean descriptors).

\section{Discussion}

% Just a few notes for limitations (you might already knew them really well, just wanted to write it somewhere):
% - does not take into account fitness variance optimization
% 

In this work, we introduced \name, a plug-and-play algorithm designed to (1) enhance the performance and reproducibility of the solutions present in a QD archive, and (2) use those solutions to find new ones in the empty cells of the corrected archive.
In the three tasks under study, we have shown that \name improves the overall performance, descriptor variance and coverage of corrected archives by a significant margin.
We also demonstrate that \name can indifferently be applied on top of multiple QD algorithms. %MAYBE? despite its performance being dependent on the performance of the initial archive provided.

However, our work has a few limitations.
First, it requires a high number of evaluations; other optimization methods could be investigated to improve the sample-efficiency.
Also, the way the cells are selected in the "archive completion" phase could be improved; for now, the cells are selected without taking into account their solution's performance or reproducibility.
Finally, we only considered reproducibility with respect to the descriptor; it would be relevant to study how to also minimize the fitness variance.
% Finally, in this work we only considered reproducibility with respect to the descriptor; it would be relevant to study how \name could also minimize the fitness variance, as expressed in the Uncertain QD setting~\cite{TEC-Manon}.

\begin{acks}
This work was supported by the Engineering and Physical Sciences Research Council (EPSRC) grant EP/V006673/1 project REcoVER.
\end{acks}

\bibliographystyle{ACM-Reference-Format}
\bibliography{sample-base}

% \newpage\hbox{}\thispagestyle{empty}\newpage
\newpage

\appendix

\section{Mathematical Justifications}

\label{appendix:proofs}

This section aims at providing mathematical proofs to justify the methods chosen in this paper.

\begin{lemma}
\label{lemma:varsum}

If $(d^{(j)})$ refer to the components of a vector of random variables $\vec d$, then:
\begin{align*}
    \sum_{j} \var{}{d^{(j)}} = \expect{}{\normeucl{ \bd - \expect{}{\bd} }^2}
\end{align*}

\end{lemma}

\begin{proof}
\begin{align*}
        \expect{}{\normeucl{ \bd - \expect{}{\bd} }^2} 
        &= \expect{}{\sum_j \left( d^{(j)} - \expect{}{d^{(j)}}\right)^2} \\
        &= \sum_j \expect{}{\left( d^{(j)} - \expect{}{d^{(j)}}\right)^2} \\
        &= \sum_{j} \var{}{d^{(j)}}
\end{align*}
\end{proof}

\begin{theorem}

If we consider a bounded descriptor space, and a cell $\cell$ such that $\expect{}{\bd}\in\cell$, then by maximizing the probability $\proba{\bd\in\cell}$ in Problem~\ref{problem:eq:proba}, we minimize an upper bound on the sum of descriptor variances $\sum_{j} \var{}{d^{(j)}}$, which are present in Problem~\ref{problem:eq:spread}.

\end{theorem}

\begin{proof}
By using Lemma~\ref{lemma:varsum}, and then the law of total expectation, we get:
\begin{align*}
    \sum_{j} \var{}{d^{(j)}} 
    &=  \expect{}{\normeucl{ \bd - \expect{}{\bd} }^2}  \tag*{see Lemma \ref{lemma:varsum}} \\
    &= \expect{}{ \normeucl{ \bd - \expect{}{\bd} }^2 | \bd\in\cell }\proba{\bd\in\cell} \\
    &\quad+ \expect{}{ \normeucl{ \bd - \expect{}{\bd} }^2 | \bd\notin\cell }\proba{\bd\notin\cell} \\
    &= \expect{}{ \normeucl{ \bd - \expect{}{\bd} }^2 | \bd\in\cell }\proba{\bd\in\cell} \\
    &\quad+ \expect{}{ \normeucl{ \bd - \expect{}{\bd} }^2 | \bd\notin\cell }(1-\proba{\bd\in\cell})
\end{align*}

Also $ \proba{\bd\in\cell} \leq 1$.
%
% Furthermore, as the cell $\cell$ is convex, then $\expect{}{\bd | \bd\in\cell}\in\cell$
%
And if we write $d_\cell$ the maximum between two points of the cell $\cell$, then, as $\expect{}{\bd} \in \cell$, then:
\begin{align*}
    \forall \bd \in\cell\; \normeucl{ \bd - \expect{}{\bd}  }^2 \leq d_{\cell}^2
\end{align*}
Thus:
\begin{align*}
    \expect{}{\normeucl{ \bd - \expect{}{\bd} }^2 | \bd\in\cell } \leq d_{\cell}^2
\end{align*}
Furthermore, the descriptor space is bounded, so, there exist a constant $\alpha$ such that
\begin{align*}
    \expect{}{\normeucl{ \bd - \expect{}{\bd} }^2 | \bd\notin\cell } \leq \alpha
\end{align*}

In the end, we obtain the following inequality:
\begin{align*}
    \sum_{j} \var{}{d^{(j)}} 
    &\leq d_\cell^2 + \alpha(1-\proba{\bd\in\cell})
\end{align*}
\fbox{%
	\parbox{\linewidth}{%
So, under the assumptions expressed in the theorem, by maximizing $\proba{\bd\in\cell}$, we minimize an upper bound on the sum of descriptor variances.
}}

\end{proof}

\begin{theorem}

If the fitness function is bounded, then by maximizing the expectation of $\noisyobj_\cell$, we also maximize a lower bound on the two objectives of Problem~\ref{problem:eq:proba}: $\proba{\bd\in\cell}$ and the expected fitness $\expect{}{f}$.

\end{theorem}

\begin{proof}

Using the law of total expectation, and the definition of $\noisyobj$ (see Equation~\ref{eq:objective:maximize}):
\begin{align*}
    \expect{}{\noisyobj} 
    &= \expect{}{\noisyobj | \bd\in\cell}\proba{\bd\in\cell} \\
    &\quad + \expect{}{\noisyobj | \bd\notin\cell} \left( 1 - \proba{\bd\in\cell} \right) \\
    &= \expect{}{\fitness | \bd\in\cell}\proba{\bd\in\cell} \\
    &\quad + \expect{}{\fitness_{min} - \normeucl{\bd-\centroid} | \bd\notin\cell} \left( 1 - \proba{\bd\in\cell} \right) \\
    &\geq \expect{}{\fitness | \bd\in\cell}\proba{\bd\in\cell} + \fitness_{min} \left( 1 - \proba{\bd\in\cell} \right) \\
    &\geq (\expect{}{\fitness | \bd\in\cell} - \fitness_{min})\proba{\bd\in\cell} + \fitness_{min} \\
\end{align*}

Thus:
\begin{align*}
    \proba{\bd\in\cell} &\geq \frac{\expect{}{\noisyobj} - \fitness_{min}}{\expect{}{\fitness | \bd\in\cell} - \fitness_{min}} \\
    &\geq \frac{\expect{}{\noisyobj} - \fitness_{min}}{\fitness_{max} - \fitness_{min}} \\
\end{align*}
\fbox{%
	\parbox{\linewidth}{%
which means that by maximizing $\expect{}{\noisyobj}$, we maximize a lower bound on $\proba{\bd\in\cell}$.
}%
}

Also, if by using the law of total expectations on the expected fitness:
\begin{align*}
    \expect{}{\fitness} 
    &= \expect{}{\fitness | \bd\in\cell}\proba{\bd\in\cell} \\
    &\quad + \expect{}{\fitness | \bd\notin\cell} \left( 1 - \proba{\bd\in\cell} \right) \\
\end{align*}
And in particular, we know that:
\begin{align*}
    \expect{}{\fitness | \bd\notin\cell} \geq  \expect{}{\fitness_{min} - \normeucl{\bd-\centroid} | \bd\notin\cell}
\end{align*}

And we have shown above that:
\begin{align*}
    \expect{}{\noisyobj} 
    &= \expect{}{\fitness | \bd\in\cell}\proba{\bd\in\cell} \\
    &\quad + \expect{}{\fitness_{min} - \normeucl{\bd-\centroid} | \bd\notin\cell} \left( 1 - \proba{\bd\in\cell} \right) \\
\end{align*}

Thus:
\begin{align*}
    \expect{}{\fitness} \geq \expect{}{\noisyobj}
\end{align*}
\fbox{%
	\parbox{\linewidth}{%
		which means that by maximizing $\expect{}{\noisyobj}$, we maximize a lower bound on the expected fitness $\expect{}{\fitness}$.
	}%
}

\end{proof}

% \newpage

\section{Archives}

\begin{figure*}[t]
    \centering
    \includegraphics[width=0.99\textwidth]{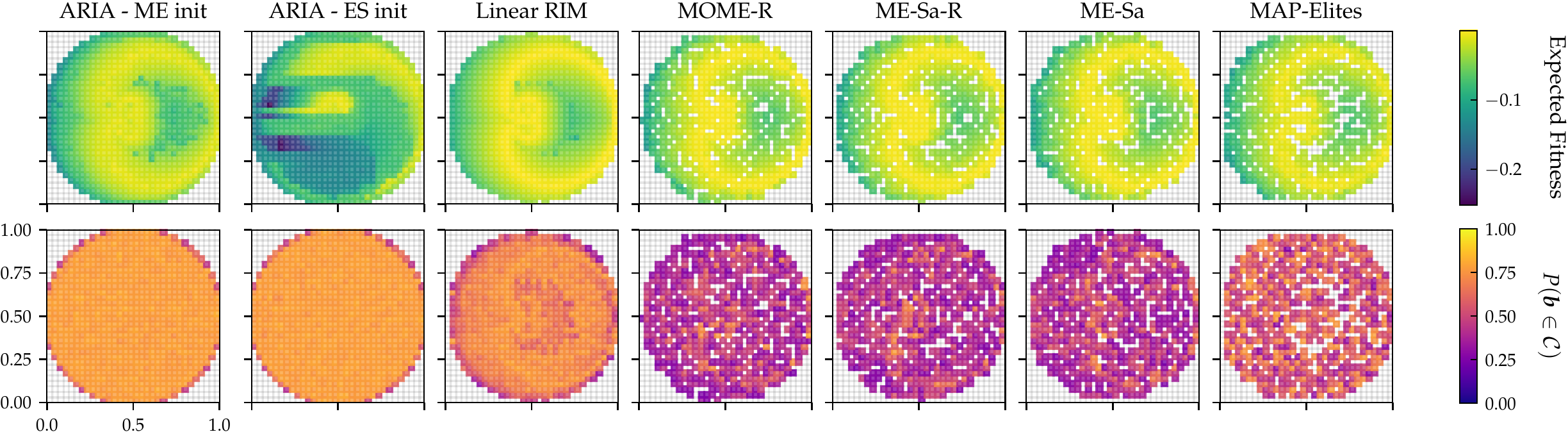}
    \caption{Distribution of the expected fitness, \nvs score, and probability of belonging to cell in an archive returned by each algorithm on the Arm task.
For each algorithm, the chosen archive is the one achieving the median QD-Score out of 10 replications.
As the Arm task is not a QD-RL task, \pgame can not directly be applied on it.
Also, the variance plots are not shown, as on the Arm task, all solutions have the same descriptor variance (see Formula~\ref{eq:armfitdesc}).
}
    \label{fig:repertoires_arm}
\end{figure*}

\begin{figure*}[t]
    \centering
    \includegraphics[width=0.99\textwidth]{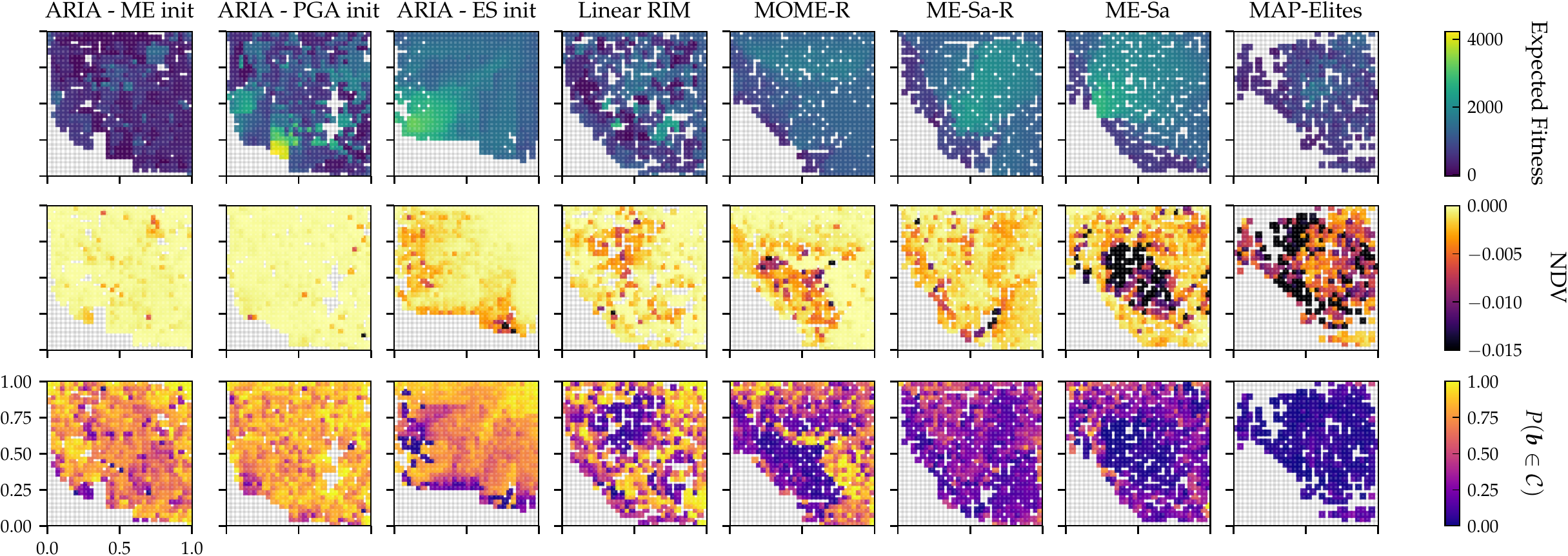}
    \caption{Distribution of the expected fitness, \nvs score, and probability of belonging to cell in an archive returned by each algorithm on the Walker task.
For each algorithm, the chosen archive is the one achieving the median QD-Score out of 10 replications.}
    \label{fig:repertoires_walker}
\end{figure*}

We provide on Figure~\ref{fig:repertoires_arm} and Figure~\ref{fig:repertoires_walker} visualizations of the archives returned by all algorithms respectively for the Arm and Walker tasks.
These figures complete the visualizations provided in Figure~\ref{fig:repertoires_ant} for the Ant task.

\section{Walker Corrected Archive Analysis}

\label{appendix:extendedanalysiswalker}

\begin{figure*}[t]
    \centering
    \includegraphics[width=0.5\textwidth]{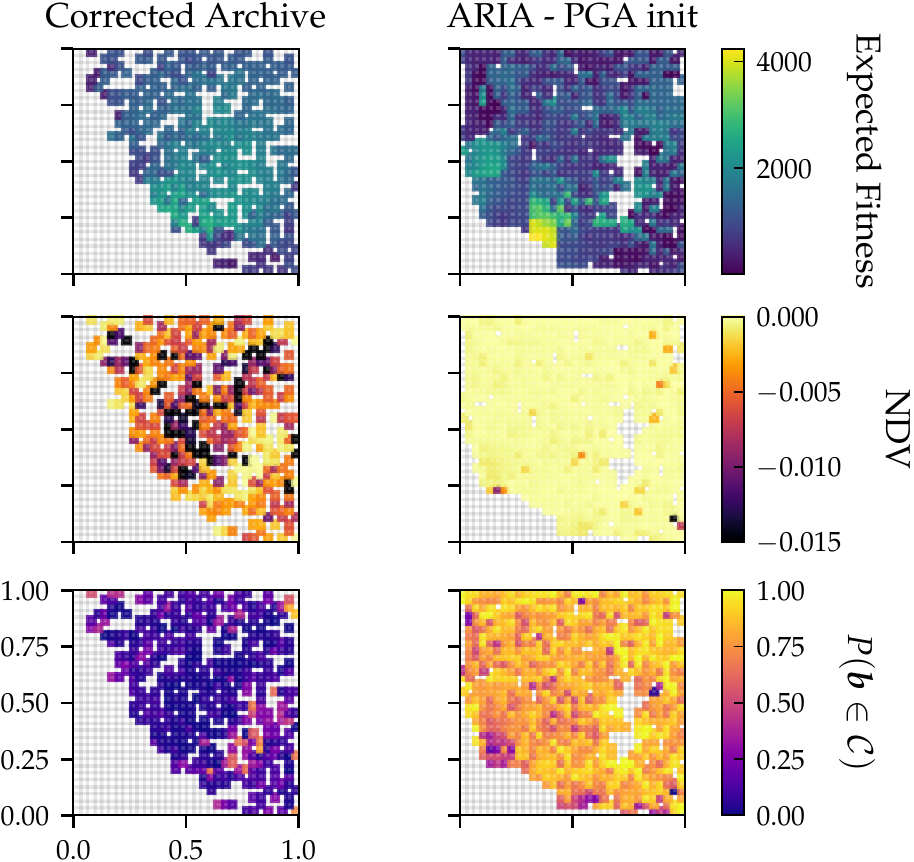}
    \caption{Distribution of the expected fitness, \nvs score, and probability of belonging to cell for the corrected archive and \namePGA archive from the Walker Task on Figure~\ref{fig:p_01_e_02_improvement_ant_walker}.}
    \label{fig:supplwalkerarchive}
\end{figure*}

We present on Figure~\ref{fig:supplwalkerarchive}, the Negative Descriptor Variance (NDV) scores and the probability values for the corrected archive and the \namePGA archive.
We notice that the archive obtained by \namePGA loses some expected fitness in different cells.
Nonetheless, \namePGA primarily optimizes the reproducibility before optimizing the performance (see Formula~\ref{eq:objective:maximize}), and that is confirmed by the NDV and probability values in Figure~\ref{fig:supplwalkerarchive}.
Interestingly, some holes appear on the \namePGA archive, at cells which were occupied in the Corrected Archive; this means the Reproducibility Improvement Mechanism may require some additional fine-tuning.

\end{document}